\documentclass[11pt,a4paper]{article}

\usepackage[hyperref]{acl2021}
\usepackage{times}
\usepackage{latexsym}

\usepackage{microtype}

\aclfinalcopy 
\def\aclpaperid{874} 

\usepackage{amsmath}
\usepackage{amssymb}

\usepackage{caption}
\usepackage{subcaption}
\usepackage{comment}

\usepackage{pifont}

\usepackage{paralist}
\usepackage{mathtools}
\usepackage{multirow}
\usepackage[capitalise]{cleveref}
\usepackage{booktabs}
\usepackage{svg}
\usepackage[english]{babel}
\usepackage{blindtext}

\usepackage{glossaries}
\glsdisablehyper

\newacronym{ODQA}{ODQA}{Open-Domain Question Answering}
\newacronym{MRS}{MRS}{Machine Reading at Scale}
\newacronym{ACT}{ACT}{Adaptive Computation Time}
\newacronym{AC}{AC}{Adaptive Computation}
\newacronym{UT}{UT}{Universal Transformers}
\newacronym{APE}{APE}{Adaptive Passage Encoder}

\title{Training Adaptive Computation for Open-Domain Question Answering with Computational Constraints}

\author{
  Yuxiang Wu \quad Pasquale Minervini \quad Pontus Stenetorp \quad Sebastian Riedel \\
  University College London \\
  { \normalsize \tt 
  \{yuxiang.wu,p.minervini,p.stenetorp,s.riedel\}@cs.ucl.ac.uk}
}

\date{}

\begin{document}
\maketitle
\begin{abstract}

Adaptive Computation (\acrshort{AC}) has been shown to be effective in improving the efficiency of Open-Domain Question Answering (\acrshort{ODQA}) systems. 
However, current AC approaches require tuning of all model parameters, and 
training state-of-the-art \acrshort{ODQA} models 
requires significant computational resources that may not be available for most researchers.
%
We propose \emph{Adaptive Passage Encoder}, an \acrshort{AC} method that can be applied to an existing \acrshort{ODQA} model and can be trained efficiently on a single GPU.
It keeps the parameters of the base \acrshort{ODQA} model fixed, but it overrides the default layer-by-layer computation of the encoder with an \acrshort{AC} policy that is trained to optimise the computational efficiency of the model.
Our experimental results show that our method improves upon a state-of-the-art model on two datasets,
and is also more accurate than previous \acrshort{AC} methods due to the stronger base \acrshort{ODQA} model.
All source code and datasets are available at \url{https://github.com/uclnlp/APE}.

\end{abstract}

\section{Introduction}

Open-Domain Question Answering (\acrshort{ODQA}) requires finding relevant information for a given question and aggregating the information to produce an answer.
The retriever-reader architecture, popularised by~\citet{DBLP:conf/acl/ChenFWB17}, has shown great success in this task. 
The retriever acquires a set of documents from external sources (e.g., Wikipedia) and the reader extracts the answer spans from these documents~\citep{DBLP:conf/acl/GardnerC18,DBLP:conf/naacl/YangXLLTXLL19, DBLP:conf/emnlp/WangNMNX19, DBLP:conf/emnlp/MinCHZ19,DBLP:conf/iclr/AsaiHHSX20}.
%
%
%
%
Recently, \citet{AmbigQA, RAG, FiD}
showed that generative reader models that exploit an encoder-decoder architecture can significantly outperform previous extractive models, thanks to their better capability in aggregating and combining evidence from multiple passages.
However, these generative models are much more computationally expensive than extractive models,
and often need to be trained with a large number of passages,
making it hard to train these models for most researchers~\citep{DBLP:journals/cacm/SchwartzDSE20}. 


\begin{figure}[t]
\begin{center}
\includegraphics[width=\columnwidth]{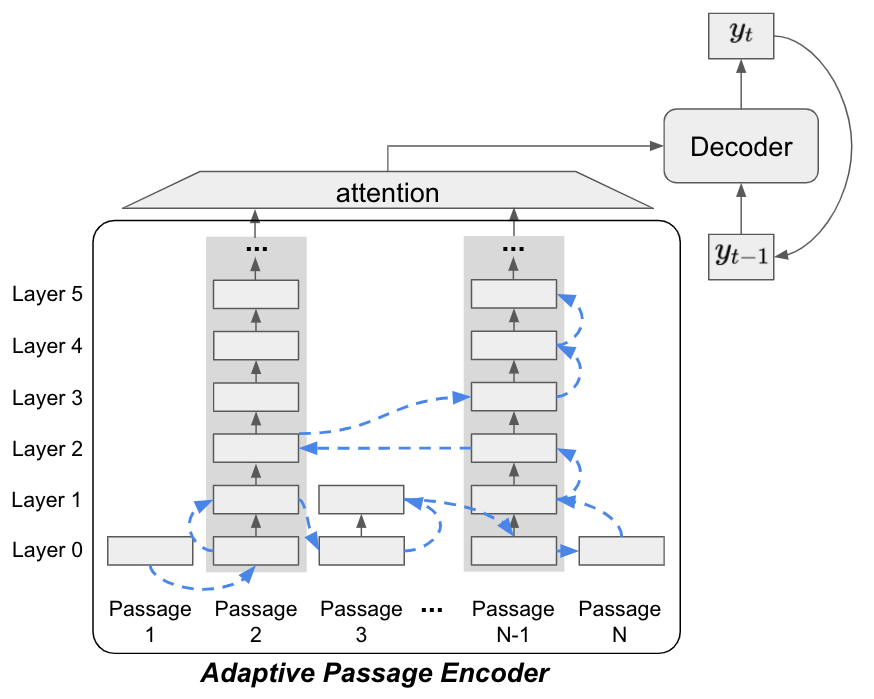}
\end{center}
\caption{Overview of our approach. The adaptive passage encoder overrides the layer-by-layer computation of the encoder with an adaptive computation policy (indicated in blue dash arrows).}
\label{fig:splash}
\end{figure}

%
\citet{SkylineBuilder} show that Adaptive Computation (\acrshort{AC}) can significantly improve the efficiency of extractive \acrshort{ODQA} models at inference time.
However, it requires fine-tuning all model parameters with a multitask learning objective,
making it computationally challenging to apply this method to current state-of-the-art models.

In this work, we explore an efficient approach to apply adaptive computation to large generative \acrshort{ODQA} models.
We introduce the \emph{Adaptive Passage Encoder} (\acrshort{APE}), a module that can be added to the encoder of an existing \acrshort{ODQA} model, which has the following features:
\begin{inparaenum}[\itshape 1\upshape)]
\item{it efficiently reuses the encoder's hidden representations for calculating the \acrshort{AC} priorities;}
\item{it does not require tuning of the base model and hence allows efficient training under limited resource;}
\item{it does not require confidence calibration.}
%
\end{inparaenum}
%
%
%
Our experimental results on NaturalQuestions and TriviaQA show that our method improves the performance of the state-of-the-art model FiD~\citep{FiD},
while also producing more accurate results (12.4\% EM) than the \acrshort{AC} method proposed by \citet{SkylineBuilder}. 
%
%
%


\section{Related Work}


\paragraph{Open Domain Question Answering}
\acrshort{ODQA} is a task that aims to answer a factoid question given a document corpus.
Most works in this domain follow a \emph{retriever}-\emph{reader} design first proposed by \citet{DBLP:conf/acl/ChenFWB17}. 
The retriever collects a set of relevant passages, then the reader comprehends and aggregates the information from multiple passages to produce the answer.
%
%
Depending on the design of the reader model, these systems could be further categorised into \emph{extractive models} and \emph{generative models}.
Extractive models~\citep{DBLP:conf/emnlp/MinCHZ19, DBLP:conf/naacl/YangXLLTXLL19, DBLP:conf/emnlp/WangNMNX19, DBLP:conf/iclr/AsaiHHSX20, DPR} exploit an answer extraction model to predict the probabilities of answer spans, and use global normalisation~\citep{DBLP:conf/acl/GardnerC18} to aggregate the answer probabilities across multiple passages.

However, thanks to recent advances in sequence-to-sequence pretrained language models~\citep{T5,BART}, generative \acrshort{ODQA} models~\citep{AmbigQA,RAG,FiD} achieve significant improvement upon extractive models, demonstrating stronger capability in combining evidence from multiple passages.
We focus on generative models in this work.
%
%
%
%
%
%
%
%


\paragraph{Passage Retrieval and Re-Ranking} 
Passage retrievers in \acrshort{ODQA} systems are initially based on sparse vector representations.
\citet{DBLP:conf/acl/ChenFWB17} use TF-IDF, whereas \citet{DBLP:conf/naacl/YangXLLTXLL19, DPR,DBLP:conf/emnlp/WangNMNX19} rely on BM25 for ranking passages~\citep{DBLP:journals/jd/Robertson04}.
Recently, \citet{DPR,RAG,GautierTeacher} achieved substantial increase in retrieval performance using dense representations.
Our work is based on the retrieval results from a dense retriever~\citep{FiD}, but we show that the proposed method can still improve the quality of the support passages despite the strong retrieval performance.

\citet{DBLP:journals/corr/abs-1901-04085, DBLP:journals/corr/abs-1904-07531, DBLP:journals/corr/abs-2101-00294} show that adding a separate cross-encoder re-ranker can improve the performance, but that comes with a significant increase of the computation at train or inference time.
Despite that our proposed adaptive passage encoder can be viewed as an encoder with an integrated re-ranker, the focus of our work is to improve the computational efficiency, namely, enhancing the performance without a substantial increase in computation.
%


\paragraph{Adaptive Computation}
Adaptive computation allows the model to condition the computation cost on the input. 
For example, \citet{DBLP:conf/acl/SchwartzSSDS20,DBLP:conf/acl/LiuZWZDJ20,DeeBert} propose models that can dynamically decide to early exit at intermediate layers when the confidence at the layer exceeds a threshold. They show that adaptively early exiting can significantly reduce the computational cost for various sequence classification tasks.
Closest to our work, \citet{SkylineBuilder} introduced adaptive computation for extractive \acrshort{ODQA} models. We extend adaptive computation to generative \acrshort{ODQA} models, and our approach can be incorporated in existing generative \acrshort{ODQA} models without finetuning the base model.

\newcommand{\layer}{\ensuremath{\mathrm{TransformerLayer}}}
\newcommand{\outputlayer}{\ensuremath{\mathrm{OutputLayer}}}
\newcommand{\logitslayer}{\ensuremath{\mathrm{Logits}}}
\newcommand{\softmax}{\ensuremath{\mathrm{Softmax}}}
\newcommand{\sigmoid}{\ensuremath{\mathrm{Sigmoid}}}
\newcommand{\mlp}{\ensuremath{\mathrm{MLP}}}

\newcommand{\problayer}{\ensuremath{\mathrm{P}_{\text{out}}}}
\newcommand{\logits}{\ensuremath{\mathbf{s}}}
\newcommand{\x}{\ensuremath{\mathbf{x}}}
\newcommand{\threshold}{\ensuremath{\tau}}
\newcommand{\noanswer}{\ensuremath{\mathrm{NoAnswer}}}
\newcommand{\hasanswer}{\ensuremath{\mathrm{HasAnswer}}}

\newcommand{\hidden}{\ensuremath{\mathbf{a}}}
\newcommand{\hiddent}{\ensuremath{\mathbf{h}}}

\newcommand{\result}{\ensuremath{\mathbf{y}}}
\newcommand{\query}{\ensuremath{\mathbf{q}}}
\newcommand{\corpus}{\ensuremath{C}}
\newcommand{\docset}{\ensuremath{D}}
\newcommand{\reader}{\ensuremath{\mathrm{Reader}}}
\newcommand{\passagereader}{\ensuremath{\mathrm{PReader}}}
\newcommand{\aggregate}{\ensuremath{\mathrm{Agg}}}
\newcommand{\CLS}{\ensuremath{\mathrm{CLS}}}
\newcommand{\currepr}{\ensuremath{A}}
\newcommand{\curheight}{\ensuremath{H}}
\newcommand{\state}{\ensuremath{S}}
\newcommand{\height}{\ensuremath{h}}
\newcommand{\buildupphase}{\ensuremath{\mathrm{BuildUp}}}
\newcommand{\outputphase}{\ensuremath{\mathrm{Output}}}

\newcommand{\numoutputs}{\ensuremath{m}}
\newcommand{\priority}{\ensuremath{p}}
\newcommand{\heightembed}{\ensuremath{\mathrm{HeightEmb}}}
\newcommand{\rankembed}{\ensuremath{\mathrm{IndexEmb}}}
\newcommand{\action}{\ensuremath{\mathbf{i}}}
\newcommand{\prioritymodel}{\ensuremath{\mathrm{PriorityModel}}}
\newcommand{\alllayer}{\ensuremath{\mathrm{AnyLayer}}}
\newcommand{\lastlayer}{\ensuremath{\mathrm{LastLayer}}}

\newcommand{\devzero}{\ensuremath{\mathrm{dev_{0}}}}
\newcommand{\devone}{\ensuremath{\mathrm{dev_{1}}}}

\newcommand{\numpassages}{\ensuremath{n}}  
\newcommand{\budget}{\ensuremath{b}}  

\section{Method}
In this section, we will introduce the base model
and how our proposed adaptive passage encoder works with it.

\subsection{Base Model}
Large generative \acrshort{ODQA} models~\citep{RAG,FiD} share a similar encoder-decoder architecture.
They first concatenate the question with all retrieved passages.
Then the encoder encodes all passages and produces their hidden representations $h_1^L, \cdots, h_N^L$, where $L$ is the number of encoder layers and $N$ is the number of retrieved passages.
We denote the hidden representation of the $i$-th passage at its $j$-th encoder layer as $h_i^j$.
The decoder will attend to these hidden representations and generate the answer tokens sequentially.


\subsection{Adaptive Passage Encoder}
As shown in \cref{fig:splash}, the adaptive passage encoder overrides the layer-by-layer computation of the encoder of the base model with an adaptive computation policy.
It adds two components on top of the base encoder to define the policy: an answerability prediction model $\hasanswer$ and a scheduler.

The $\hasanswer$ model predicts the probability that a passage contains an answer to the question, given its hidden representation $h_i^j$. 
It first pools hidden representation $h_i^j$ into a vector, then feeds the pooled representation to a multi-layer perceptron to produce the probability $p_i^j$.

The scheduler is then responsible for the selection and prioritisation of passages that are likely to contain the answer~\citep{SkylineBuilder}.
As shown by the blue arrows in \cref{fig:splash}, the scheduler learns a scheduling policy to allocate encoder layer computation to passages.
The scheduler will exit in early layers for those spurious passages while allocating more layers to the ones that it finds promising.
%
%

To achieve this goal, the scheduler produces a \emph{priority} score $q_n$ for each passage:
\begin{equation}
    q_n = \sigma(g(p_n^{l_n}, n, {l_n}))  p_n^{l_n} + f(p_n^{l_n}, n, {l_n})
    \label{eq:priority}
\end{equation}
where $n$ is the passage rank by the retriever, $l_n$ is the index of its current encoder layer,
$g$ and $f$ are two multi-layer perceptrons that learn the weight and bias respectively.
Starting at the initial layer for all passages, the scheduler will select a passage with the maximum priority, forward one encoder layer for it $l'_n = l_n + 1$, and updates its priorities $q_n$ with its new hidden representation $h_n^{l'_n}$ and has-answer probability $p_n^{l'_n}$.
This process will iterate for $B$ (budget) steps,
and only $k$ passages with the most layers computed are retained in the end.

\subsection{Training the \acrlong{APE}}

Differently from \citet{SkylineBuilder}, our method does not require tuning the underlying base model.
%
Since the number of parameters introduced by the $\hasanswer$ model and the scheduler is less than 4\% of the base model, \acrshort{APE} can be trained very efficiently.
The $\hasanswer$ model is first trained with cross-entropy loss, supervised by the has-answer labels of the passages.
Then we fix $\hasanswer$ and train the scheduler with REINFORCE algorithm~\citep{Williams:92} to maximise the expected return, which is defined to encourage selection and prioritisation of passages that contain the answer.
The selection action gains a positive reward $(1-c)$ if it selects a relevant passage, otherwise a negative reward $- c$.
%
Since the weight $g$ and bias $f$ in \cref{eq:priority} are automatically learned during the training of the scheduler, our method does not require confidence calibration of the $\hasanswer$ model, unlike the method proposed by \citet{SkylineBuilder}.

\section{Experiments} \label{sec:exp}

\subsection{Experimental Setup}

\paragraph{Datasets}

Following~\cite{DBLP:conf/acl/LeeCT19,FiD}, we evaluate our method on NaturalQuestions~\citep{NaturalQuestions} and TriviaQA~\citep{TriviaQA} whose statistics are shown in \cref{tab:datasets}.

\begin{table}[t]
\begin{center}
\resizebox{\columnwidth}{!}{
    \begin{tabular}{lccc}
      \toprule
       & {\bf Train} \ & {\bf Validation} \ & {\bf Test} \\
      \midrule
      {NaturalQuestions} & 79,168 & 8,757 & 3,610  \\
      {TriviaQA} & 78,785 & 8,837 & 11,313 \\
      \bottomrule
    \end{tabular}
}
\caption{Number of samples of the evaluated datasets.} \label{tab:datasets}
\end{center}
\end{table}

\paragraph{Evaluation Metrics}

Following~\citet{SkylineBuilder},
we conduct the evaluation under different computational costs at inference time.
Since the number of passages $k$ is almost linearly correlated with memory consumption and number of operations, we evaluate the performances with various number of passages $k \in \{5,10,20\}$.
%
%
To evaluate the end performance of \acrshort{ODQA} models, we use the standard Exact Match (EM) score, which is the proportion of questions whose predicted answer matches exactly with the ground truth.
We also include the unrestricted setting to compare the best performances of different models.

\paragraph{Technical Details}
We use FiD~\citep{FiD} as our base model. FiD-base and FiD-large contain $L=12$ and $24$ layers respectively, and we set the budget $B=Lk$.
For the pooling operation in the $\hasanswer$ model, we found max-pooling works better than mean-pooling and the [CLS] token, so max-pooling is used in all our experiments.
We use discount factor $\gamma=0.8$ and step penalty $c=0.1$ during the REINFORCE training of the scheduler.
%
More hyperparameters are presented in \cref{sec:appendix_hparams}.

\begin{table*}[hbt]
%
\begin{center}
\resizebox{\textwidth}{!}{
\begin{tabular}{lllllllll}
  \toprule
  \  &  \multicolumn{4}{c}{\bf NaturalQuestions} \ & \multicolumn{4}{c}{\bf TriviaQA}  \\
  \midrule
  & Top-5 & Top-10 & Top-20 & Unrestricted &  Top-5  & Top-10  & Top-20 &  Unrestricted \\
  \midrule
  SkylineBuilder~\citep{SkylineBuilder} & 34.4 & 34.2 & - & 34.2 & - & - & - & - \\ 
  DPR~\citep{DPR} & - & 40.8 & - & 41.5 & - & - & - & 57.9 \\
  DPR (our implementation) & 38.4 & 40.2 & 40.2 & 40.2 & - & - & - & - \\
  \midrule
  RAG~\citep{RAG} & \textbf{43.5} & 44.1 & 44.1 & 44.5 & - & - & - & 56.1 \\
  \midrule
  FiD-base~\citep{FiD} & 39.5 & 42.9 & 45.3 & 48.2 & 53.9 & 57.9 & 60.7 & 65.0 \\
  Ours (\acrshort{APE}+FiD-base) & \textbf{40.3} & \textbf{43.7} & \textbf{46.0} & 48.2 & \textbf{55.4}\mbox{*} & \textbf{59.0}\mbox{*} & \textbf{62.0}\mbox{*} & 65.0 \\
  \midrule
  FiD-large~\citep{FiD} & 42.5 & 45.8 & 48.3 & 51.4 & 57.2 & 60.6 & 63.7 & 67.6 \\
  Ours (\acrshort{APE}+FiD-large) & \textbf{43.4} & \bf{46.6} & \bf{49.1} & 51.4 & \textbf{57.9} & \textbf{61.4}\mbox{*} & \textbf{64.1}\mbox{*} & 67.6 \\
  \bottomrule
\end{tabular}
}
\caption{Exact match scores on NaturalQuestions and TriviaQA test sets. \mbox{*} indicates statistical significance.} \label{tab:em}

\end{center}
\end{table*}
\begin{table*}[t]
%
\begin{center}
\resizebox{\textwidth}{!}{
\begin{tabular}{lllllllll}
  \toprule
  \  &  \multicolumn{4}{c}{\bf NaturalQuestions} \ & \multicolumn{4}{c}{\bf TriviaQA}  \\
  \midrule
   & Top-5 & Top-10  & Top-20 & Top-100 & Top-5  & Top-10 & Top-20 & Top-100 \\
  \midrule
  BM25~\citep{DBLP:conf/acl/LeeCT19} & - & - & 59.1 & 73.7 & - & - & 66.9 & 76.7 \\
  DPR~\citep{DPR} & 67.1 & - & 78.4 & 85.4 & - & - & 79.4 & 85.0 \\
  \midrule
  FiD~\citep{FiD} & 66.2 & 73.9 & 79.2 & 86.1 & 69.8 & 74.9 & 78.9 & 84.8 \\
  Ours (\acrshort{APE}+FiD-base) & \textbf{67.4}\mbox{*} &  75.1\mbox{*}  &  \textbf{80.4}\mbox{*} & 86.1  & \textbf{70.8}\mbox{*}  &  \textbf{75.8}\mbox{*}  & \textbf{79.5}  & 84.8  \\
  Ours (\acrshort{APE}+FiD-large) & 67.2  &  \textbf{75.4}\mbox{*}  &  80.2\mbox{*} & 86.1  & 70.4  & 75.6\mbox{*} &  79.2 &  84.8 \\
  \bottomrule
\end{tabular}
}
\caption{Top-k retrieval accuracy scores on NaturalQuestions and TriviaQA test sets. \mbox{*} indicates statistical significance.} \label{tab:retrieval-top-k}
\end{center}
\end{table*}

\paragraph{Computational Feasibility}  Tuning a FiD-base model with $k=20$ or a FiD-large model with $k=10$ (batch size=1) would yield out-of-memory errors on a V100 (16GB) GPU. Hence, it is infeasible to train FiD with the previous AC method~\citep{SkylineBuilder} in our setting. However, training with our proposed approach can be done in the same setting with a batch size 4 or larger within 8-15 hours.

\subsection{Experimental Results}

As shown in \cref{tab:em} under restricted top-$k$, our proposed method improves upon the FiD model on both datasets, and by a statistically significant margin on TriviaQA.
It also outperforms the previous AC method~\citep{SkylineBuilder} by 12.4\% when $k=10$ due to the stronger base model.
The addition of \acrshort{APE} allows FiD to significantly outperform RAG~\citep{RAG} on NaturalQuestions when $k \in \{ 10, 20 \}$.

Previous adaptive computation methods~\citep{SkylineBuilder,DBLP:conf/acl/SchwartzSSDS20} was reported to have plateaued or degraded performances in the unrestricted setting.
However, \cref{tab:em} shows that our approach does not have this issue.

\subsection{Analysis of Passage Quality}

To understand how \acrshort{APE} outperforms the baselines, we analyse the quality of the final top-$k$ passages retained by \acrshort{APE}. \cref{tab:retrieval-top-k} reports the top-$k$ retrieval accuracy of the top-$k$ passages.
The results show that the top-$k$ accuracy of the selected collection of documents by \acrshort{APE} is significantly better than BM25, DPR, and FiD, which are strong retrieval baselines for \acrshort{ODQA}.
Combined with \cref{tab:em}, it indicates that the better passage quality yielded by \acrshort{APE} helps to improve the end \acrshort{ODQA} performance of the model.


\section{Conclusions}
In this work, we explore an adaptive computation method that can be efficiently applied to an existing generative \acrshort{ODQA} model.
We find that, by replacing the encoder of generative \acrshort{ODQA} models with our proposed adaptive passage encoder, we can train an effective adaptive computation policy without tuning the base model.
This allows applying adaptive computation to large state-of-the-art generative models, which was previously challenging computation-wise.
Our experimental results show that our method produces more accurate results than a state-of-the-art generative model on both NaturalQuestions and TriviaQA,
and it outperforms the previous AC method by a large margin.
The analysis also shows that our approach achieves better passage quality that leads to improvements in ODQA performance.

\section*{Acknowledgments}

The first author would like to thank his wife Jane for her love and support throughout the years. We would also like to thank Gautier Izacard and Edouard Grave for their help with using FiD. 
This research was supported by the European Union's Horizon 2020 research and innovation programme under grant agreement no. 875160.

\bibliography{bibliography}

\begin{thebibliography}{26}
\expandafter\ifx\csname natexlab\endcsname\relax\def\natexlab#1{#1}\fi

\bibitem[{Asai et~al.(2020)Asai, Hashimoto, Hajishirzi, Socher, and
  Xiong}]{DBLP:conf/iclr/AsaiHHSX20}
Akari Asai, Kazuma Hashimoto, Hannaneh Hajishirzi, Richard Socher, and Caiming
  Xiong. 2020.
\newblock Learning to retrieve reasoning paths over wikipedia graph for
  question answering.
\newblock In \emph{{ICLR}}. OpenReview.net.

\bibitem[{Chen et~al.(2017)Chen, Fisch, Weston, and
  Bordes}]{DBLP:conf/acl/ChenFWB17}
Danqi Chen, Adam Fisch, Jason Weston, and Antoine Bordes. 2017.
\newblock \href {https://doi.org/10.18653/v1/P17-1171} {Reading wikipedia to
  answer open-domain questions}.
\newblock In \emph{{ACL} {(1)}}, pages 1870--1879. Association for
  Computational Linguistics.

\bibitem[{Clark and Gardner(2018)}]{DBLP:conf/acl/GardnerC18}
Christopher Clark and Matt Gardner. 2018.
\newblock \href {https://doi.org/10.18653/v1/P18-1078} {Simple and effective
  multi-paragraph reading comprehension}.
\newblock In \emph{{ACL} {(1)}}, pages 845--855. Association for Computational
  Linguistics.

\bibitem[{Izacard and Grave(2020{\natexlab{a}})}]{GautierTeacher}
Gautier Izacard and Edouard Grave. 2020{\natexlab{a}}.
\newblock Distilling knowledge from reader to retriever for question answering.
\newblock \emph{CoRR}, abs/2012.04584.

\bibitem[{Izacard and Grave(2020{\natexlab{b}})}]{FiD}
Gautier Izacard and Edouard Grave. 2020{\natexlab{b}}.
\newblock Leveraging passage retrieval with generative models for open domain
  question answering.
\newblock \emph{CoRR}, abs/2007.01282.

\bibitem[{Joshi et~al.(2017)Joshi, Choi, Weld, and Zettlemoyer}]{TriviaQA}
Mandar Joshi, Eunsol Choi, Daniel~S. Weld, and Luke Zettlemoyer. 2017.
\newblock \href {https://doi.org/10.18653/v1/P17-1147} {Triviaqa: {A} large
  scale distantly supervised challenge dataset for reading comprehension}.
\newblock In \emph{{ACL} {(1)}}, pages 1601--1611. Association for
  Computational Linguistics.

\bibitem[{Karpukhin et~al.(2020)Karpukhin, Oguz, Min, Lewis, Wu, Edunov, Chen,
  and Yih}]{DPR}
Vladimir Karpukhin, Barlas Oguz, Sewon Min, Patrick S.~H. Lewis, Ledell Wu,
  Sergey Edunov, Danqi Chen, and Wen{-}tau Yih. 2020.
\newblock \href {https://doi.org/10.18653/v1/2020.emnlp-main.550} {Dense
  passage retrieval for open-domain question answering}.
\newblock In \emph{{EMNLP} {(1)}}, pages 6769--6781. Association for
  Computational Linguistics.

\bibitem[{Kwiatkowski et~al.(2019)Kwiatkowski, Palomaki, Redfield, Collins,
  Parikh, Alberti, Epstein, Polosukhin, Devlin, Lee, Toutanova, Jones, Kelcey,
  Chang, Dai, Uszkoreit, Le, and Petrov}]{NaturalQuestions}
Tom Kwiatkowski, Jennimaria Palomaki, Olivia Redfield, Michael Collins,
  Ankur~P. Parikh, Chris Alberti, Danielle Epstein, Illia Polosukhin, Jacob
  Devlin, Kenton Lee, Kristina Toutanova, Llion Jones, Matthew Kelcey,
  Ming{-}Wei Chang, Andrew~M. Dai, Jakob Uszkoreit, Quoc Le, and Slav Petrov.
  2019.
\newblock \href {https://doi.org/10.1162/tacl_a_00276} {Natural questions: a
  benchmark for question answering research}.
\newblock \emph{Trans. Assoc. Comput. Linguistics}, 7:452--466.

\bibitem[{Lee et~al.(2019)Lee, Chang, and Toutanova}]{DBLP:conf/acl/LeeCT19}
Kenton Lee, Ming{-}Wei Chang, and Kristina Toutanova. 2019.
\newblock \href {https://doi.org/10.18653/v1/P19-1612} {Latent retrieval for
  weakly supervised open domain question answering}.
\newblock In \emph{{ACL} {(1)}}, pages 6086--6096. Association for
  Computational Linguistics.

\bibitem[{Lewis et~al.(2020{\natexlab{a}})Lewis, Liu, Goyal, Ghazvininejad,
  Mohamed, Levy, Stoyanov, and Zettlemoyer}]{BART}
Mike Lewis, Yinhan Liu, Naman Goyal, Marjan Ghazvininejad, Abdelrahman Mohamed,
  Omer Levy, Veselin Stoyanov, and Luke Zettlemoyer. 2020{\natexlab{a}}.
\newblock \href {https://doi.org/10.18653/v1/2020.acl-main.703} {{BART:}
  denoising sequence-to-sequence pre-training for natural language generation,
  translation, and comprehension}.
\newblock In \emph{{ACL}}, pages 7871--7880. Association for Computational
  Linguistics.

\bibitem[{Lewis et~al.(2020{\natexlab{b}})Lewis, Perez, Piktus, Petroni,
  Karpukhin, Goyal, K{\"{u}}ttler, Lewis, Yih, Rockt{\"{a}}schel, Riedel, and
  Kiela}]{RAG}
Patrick S.~H. Lewis, Ethan Perez, Aleksandra Piktus, Fabio Petroni, Vladimir
  Karpukhin, Naman Goyal, Heinrich K{\"{u}}ttler, Mike Lewis, Wen{-}tau Yih,
  Tim Rockt{\"{a}}schel, Sebastian Riedel, and Douwe Kiela. 2020{\natexlab{b}}.
\newblock Retrieval-augmented generation for knowledge-intensive {NLP} tasks.
\newblock In \emph{NeurIPS}.

\bibitem[{Liu et~al.(2020)Liu, Zhou, Wang, Zhao, Deng, and
  Ju}]{DBLP:conf/acl/LiuZWZDJ20}
Weijie Liu, Peng Zhou, Zhiruo Wang, Zhe Zhao, Haotang Deng, and Qi~Ju. 2020.
\newblock \href {https://doi.org/10.18653/v1/2020.acl-main.537} {Fastbert: a
  self-distilling {BERT} with adaptive inference time}.
\newblock In \emph{{ACL}}, pages 6035--6044. Association for Computational
  Linguistics.

\bibitem[{Mao et~al.(2021)Mao, He, Liu, Shen, Gao, Han, and
  Chen}]{DBLP:journals/corr/abs-2101-00294}
Yuning Mao, Pengcheng He, Xiaodong Liu, Yelong Shen, Jianfeng Gao, Jiawei Han,
  and Weizhu Chen. 2021.
\newblock Reader-guided passage reranking for open-domain question answering.
\newblock \emph{CoRR}, abs/2101.00294.

\bibitem[{Min et~al.(2019)Min, Chen, Hajishirzi, and
  Zettlemoyer}]{DBLP:conf/emnlp/MinCHZ19}
Sewon Min, Danqi Chen, Hannaneh Hajishirzi, and Luke Zettlemoyer. 2019.
\newblock \href {https://doi.org/10.18653/v1/D19-1284} {A discrete hard {EM}
  approach for weakly supervised question answering}.
\newblock In \emph{{EMNLP/IJCNLP} {(1)}}, pages 2851--2864. Association for
  Computational Linguistics.

\bibitem[{Min et~al.(2020)Min, Michael, Hajishirzi, and Zettlemoyer}]{AmbigQA}
Sewon Min, Julian Michael, Hannaneh Hajishirzi, and Luke Zettlemoyer. 2020.
\newblock \href {https://doi.org/10.18653/v1/2020.emnlp-main.466} {Ambigqa:
  Answering ambiguous open-domain questions}.
\newblock In \emph{{EMNLP} {(1)}}, pages 5783--5797. Association for
  Computational Linguistics.

\bibitem[{Nogueira and Cho(2019)}]{DBLP:journals/corr/abs-1901-04085}
Rodrigo Nogueira and Kyunghyun Cho. 2019.
\newblock Passage re-ranking with {BERT}.
\newblock \emph{CoRR}, abs/1901.04085.

\bibitem[{Qiao et~al.(2019)Qiao, Xiong, Liu, and
  Liu}]{DBLP:journals/corr/abs-1904-07531}
Yifan Qiao, Chenyan Xiong, Zhenghao Liu, and Zhiyuan Liu. 2019.
\newblock Understanding the behaviors of {BERT} in ranking.
\newblock \emph{CoRR}, abs/1904.07531.

\bibitem[{Raffel et~al.(2020)Raffel, Shazeer, Roberts, Lee, Narang, Matena,
  Zhou, Li, and Liu}]{T5}
Colin Raffel, Noam Shazeer, Adam Roberts, Katherine Lee, Sharan Narang, Michael
  Matena, Yanqi Zhou, Wei Li, and Peter~J. Liu. 2020.
\newblock Exploring the limits of transfer learning with a unified text-to-text
  transformer.
\newblock \emph{J. Mach. Learn. Res.}, 21:140:1--140:67.

\bibitem[{Robertson(2004)}]{DBLP:journals/jd/Robertson04}
Stephen Robertson. 2004.
\newblock Understanding inverse document frequency: on theoretical arguments
  for {IDF}.
\newblock \emph{Journal of Documentation}, 60(5):503--520.

\bibitem[{Schwartz et~al.(2020{\natexlab{a}})Schwartz, Dodge, Smith, and
  Etzioni}]{DBLP:journals/cacm/SchwartzDSE20}
Roy Schwartz, Jesse Dodge, Noah~A. Smith, and Oren Etzioni. 2020{\natexlab{a}}.
\newblock Green {AI}.
\newblock \emph{Commun. {ACM}}, 63(12):54--63.

\bibitem[{Schwartz et~al.(2020{\natexlab{b}})Schwartz, Stanovsky, Swayamdipta,
  Dodge, and Smith}]{DBLP:conf/acl/SchwartzSSDS20}
Roy Schwartz, Gabriel Stanovsky, Swabha Swayamdipta, Jesse Dodge, and Noah~A.
  Smith. 2020{\natexlab{b}}.
\newblock \href {https://doi.org/10.18653/v1/2020.acl-main.593} {The right tool
  for the job: Matching model and instance complexities}.
\newblock In \emph{{ACL}}, pages 6640--6651. Association for Computational
  Linguistics.

\bibitem[{Wang et~al.(2019)Wang, Ng, Ma, Nallapati, and
  Xiang}]{DBLP:conf/emnlp/WangNMNX19}
Zhiguo Wang, Patrick Ng, Xiaofei Ma, Ramesh Nallapati, and Bing Xiang. 2019.
\newblock \href {https://doi.org/10.18653/v1/D19-1599} {Multi-passage {BERT:}
  {A} globally normalized {BERT} model for open-domain question answering}.
\newblock In \emph{{EMNLP/IJCNLP} {(1)}}, pages 5877--5881. Association for
  Computational Linguistics.

\bibitem[{Williams(1992)}]{Williams:92}
R.~J. Williams. 1992.
\newblock Simple statistical gradient-following algorithms for connectionist
  reinforcement learning.
\newblock \emph{Machine Learning}, 8:229--256.

\bibitem[{Wu et~al.(2020)Wu, Riedel, Minervini, and Stenetorp}]{SkylineBuilder}
Yuxiang Wu, Sebastian Riedel, Pasquale Minervini, and Pontus Stenetorp. 2020.
\newblock \href {https://doi.org/10.18653/v1/2020.emnlp-main.244} {Don't read
  too much into it: Adaptive computation for open-domain question answering}.
\newblock In \emph{{EMNLP} {(1)}}, pages 3029--3039. Association for
  Computational Linguistics.

\bibitem[{Xin et~al.(2020)Xin, Tang, Lee, Yu, and Lin}]{DeeBert}
Ji~Xin, Raphael Tang, Jaejun Lee, Yaoliang Yu, and Jimmy Lin. 2020.
\newblock \href {https://doi.org/10.18653/v1/2020.acl-main.204} {Deebert:
  Dynamic early exiting for accelerating {BERT} inference}.
\newblock In \emph{{ACL}}, pages 2246--2251. Association for Computational
  Linguistics.

\bibitem[{Yang et~al.(2019)Yang, Xie, Lin, Li, Tan, Xiong, Li, and
  Lin}]{DBLP:conf/naacl/YangXLLTXLL19}
Wei Yang, Yuqing Xie, Aileen Lin, Xingyu Li, Luchen Tan, Kun Xiong, Ming Li,
  and Jimmy Lin. 2019.
\newblock \href {https://doi.org/10.18653/v1/N19-4013} {End-to-end open-domain
  question answering with bertserini}.
\newblock In \emph{{NAACL-HLT} (Demonstrations)}, pages 72--77. Association for
  Computational Linguistics.

\end{thebibliography}
\bibliographystyle{acl_natbib}

\clearpage

\appendix

\section{Experimental Details} \label{sec:appendix_experiments}

\subsection{Hyper-parameters}\label{sec:appendix_hparams}

\begin{table}[!hb]
\begin{center}
    \begin{tabular}{lc}
      \toprule
      {\bf Hyper-parameter} & {\bf Value} \\
      \midrule
      learning rate & 1e-4 \\
      batch size & 24 \\
      epoch & 2 \\
      optimiser & Adam \\
      Adam $\epsilon$ & 1e-6 \\
      Adam $(\beta_1, \beta_2)$ & (0.9, 0.999) \\
      max sequence length & 256 \\
      pooling & max-pooling \\
      number of passages & 5/10/20 \\
      device & Nvidia V100 \\
      \bottomrule
    \end{tabular}
\caption{Hyper-parameters for the $\hasanswer$ model training.} \label{tab:hyperparam}
\end{center}
\end{table}

\begin{table}[!hb]
\begin{center}
    \begin{tabular}{lc}
      \toprule
      {\bf Hyper-parameter} & {\bf Value} \\
      \midrule
      learning rate & 0.01 \\
      batch size & 24 \\
      epoch & 1 \\
      optimiser & Adam \\
      max number of steps & 240 \\
      step cost $c$ & 0.1 \\
      discount factor $\gamma$ & 0.8 \\
      hidden size of MLPs & 64 \\
      number of passages & 20/30/50 \\
      \bottomrule
    \end{tabular}
\caption{Hyper-parameters for scheduler model REINFORCE training.} \label{tab:hyperparamRL}
\end{center}
\end{table}






\end{document}